# A Systematic Evaluation of Trajectory Data Curation for LoRA Fine-Tuning of Code Agents

Yunze Han[1]

[1] Vdynamics GmbH, 80935 Munich, Germany
yunze.han94@gmail.com

**Abstract.** Supervised fine-tuning (SFT) of open-weight LLMs on expert agent trajectories has emerged as a prominent approach to building capable code agents without reliance on proprietary models. A central yet underexplored question is how trajectory quality and quantity jointly shape model performance. We present a systematic empirical study of trajectory data filtering for LoRA fine-tuning of Qwen2.5-Coder-7B-Instruct on the SWE-trajectory dataset (67,074 trajectories, of which 32,161 are resolved). We propose a two-axis quality scoring framework—Efficiency and Style—and evaluate it through 16 controlled experiments spanning strategy, scale, and ablation analyses. Since 7B-scale models attain near-zero SWE-bench resolve rates, we adopt cross-entropy (CE) loss on held-out trajectories as the primary metric, validated via first-action generation: CE loss and ROUGE-L are perfectly rank-correlated (Spearman $\rho = -1.00$), with limited-sample evidence supporting but not conclusively establishing this proxy. Our results reveal a scale-dependent quality–quantity trade-off: at small scales, doubling the dataset (500→1,000) yields ~12.7% CE-loss reduction whereas the TopQ–Random gap stays <1% (Mann–Whitney $p > 0.10$); at 2,000 trajectories this same gap widens to 3.6% ($p = 0.016$). Ablation further identifies error-retry rate as the dominant sub-dimension, performing comparably to the full composite ($\Delta < 0.2\%$). Together, these findings establish trajectory-level quality scoring as a viable but scale-sensitive lever for code-agent SFT and offer a proxy-validated evaluation protocol for the regime where end-to-end resolve rate is statistically infeasible.

## 1 Introduction

Large language models (LLMs) have given rise to autonomous code agents that tackle repository-level tasks through iterative tool use [1, 2]. Code agents interleave reasoning, shell commands, file edits, and tool calls, following patterns formalized by ReAct [3] and CodeAct [4]. Benchmarks such as SWE-bench [1] have shown that frontier proprietary models (GPT-4, Claude) can resolve a non-trivial fraction of real GitHub issues [2, 5].

Because closed-source models remain expensive to deploy, supervised fine-tuning (SFT) of smaller open-weight LLMs—such as Qwen2.5-Coder [6], Code Llama [7], and DeepSeek-Coder [8]—on expert trajectories has emerged as an attractive alternative. In this regime the composition of the training set is decisive: LIMA [9] showed that 1,000 curated examples elicit strong instruction-following, and AlpaGasus [10] and DEITA [11] demonstrated that automated quality scoring can match or exceed full-dataset training. Whether these principles transfer to multi-step agent trajectories remains open.

We present a systematic study of data filtering for LoRA [12] fine-tuning of a code-agent LLM, organized around four research questions: (L1) does SFT on agent trajectories improve over the base model? (L2) does resolution-based gating beat random sampling? (L3) does composite quality filtering add gains over gating alone? (L4) which sub-dimensions matter most? Because 7B models score near zero on SWE-bench Verified, we adopt held-out CE loss as our primary metric and validate it via first-action generation. Experiments use a single base model (Qwen2.5-Coder-7B-Instruct) and a single trajectory source (SWE-trajectory); generalization to other model families and corpora remains to be tested.

Our central empirical finding is that, in this regime, data quantity effects (6.5–12.7% relative loss reduction per doubling) substantially exceed quality filtering effects: the TopQ–Random gap is 0.7% at 500 samples (n.s.) but widens to 3.6% at 2,000 ($p = 0.016$), while a TopQ vs. BottomQ sanity check ($p = 0.013$) confirms framework self-consistency. We frame this widening as an emerging signal rather than a confirmed crossover, given the test-split size (200 trajectories per split) and the limited number of scale points (500/1,000/2,000).

Contributions: (i) a multi-dimensional quality scoring framework spanning Efficiency (error-retry rate, step-count ratio) and Style (action diversity, observation utilization), with two candidates discarded due to near-zero variance in the SWE-trajectory dataset; (ii) the first controlled comparison of 16 data filtering configurations for LoRA fine-tuning of code agents, spanning strategy, scale (500/1,000/2,000), and dimension ablation including a B2-only baseline; (iii) the quantity-dominates-quality finding and its scale-dependent widening at 2,000; and (iv) a tri-partite Gold/Random/Low-Q evaluation protocol, validated by 100% gradient satisfaction across 17 checkpoints and rank-correlated with first-action ROUGE-L (Spearman $\rho = -1.00$; $n = 4$; minimum achievable two-sided $p \approx 0.083$, see Section 4.2 for caveat).

The remainder of this paper is organized as follows. Section 2 reviews related work; Section 3 presents the scoring framework and experimental design; Section 4 reports results, including proxy metric validation, scaling analysis, and the B2-only baseline; Section 5 discusses implications and limitations; Section 6 concludes.

## 2 Related Work

### 2.1 LLM-based Code Agents

LLMs for repository-level software engineering were galvanized by SWE-bench [1] (2,294 GitHub issues across 12 Python repositories), which goes beyond function-level benchmarks by requiring multi-file navigation and patch generation. SWE-agent [2] showed that wrapping GPT-4 with a purpose-built agent–computer interface achieved 12.5% on the SWE-bench test set. OpenDevin/OpenHands [5] refined the agent architecture, and SWE-bench Verified [1] became the standard reporting split, with frontier proprietary models exceeding 70% and the best open-weight agents reaching 40–51% via trajectory-based fine-tuning [15, 16, 17]. ReAct [3] established the reasoning–acting paradigm, while CodeAct [4] argued for executable-Python actions. Our work studies how to train smaller open-weight models to exhibit these behaviors via SFT on expert trajectories.

### 2.2 Data Quality in Fine-Tuning

LIMA [9] first demonstrated that data quality outweighs quantity in instruction tuning, showing that 1,000 curated pairs sufficed to elicit competitive instruction-following behavior. AlpaGasus [10] used ChatGPT as a quality oracle, outperforming full-dataset training; the IFD metric [13] scored examples by perplexity discrepancy. DEITA [11] further combined complexity and response quality with GPT-4 scoring. All of these methods were developed for general-purpose, single-turn data.

Beyond instruction tuning, large-scale data engineering has driven code-LLM progress, exemplified by Qwen2.5-Coder [6] (quality filtering, deduplication, and domain balancing across 5.5T tokens). While such work targets the token or snippet level, our study targets the trajectory level, where each example is a multi-step agent rollout. The most directly related line is SWE-agent fine-tuning: SWE-Gym [15] (32B model, <500 successful trajectories, +14% on SWE-bench Verified), SWE-smith [16] (50,000+ synthetic instances across 128 repositories, 40.2%), and R2E-Gym [17] (8,100 procedural environments with hybrid verifiers, 51%). All rely on binary outcome filtering—retaining trajectories only if the final patch passes the test suite—and conduct no systematic quality–quantity ablations at small data scales. To our knowledge, no prior work has applied multi-dimensional, continuous quality scoring to agent trajectory data, nor characterized how such criteria trade off with dataset size in the agentic fine-tuning regime.

### 2.3 Parameter-Efficient Fine-Tuning

LoRA [12] injects trainable low-rank decomposition matrices into attention weights, reducing trainable parameters by orders of magnitude; QLoRA [14] further combines 4-bit quantization with LoRA to enable fine-tuning of large models on a single consumer GPU. LoRA's efficiency permits the rapid iteration our 16-configuration matrix requires.

# 3 Methodology

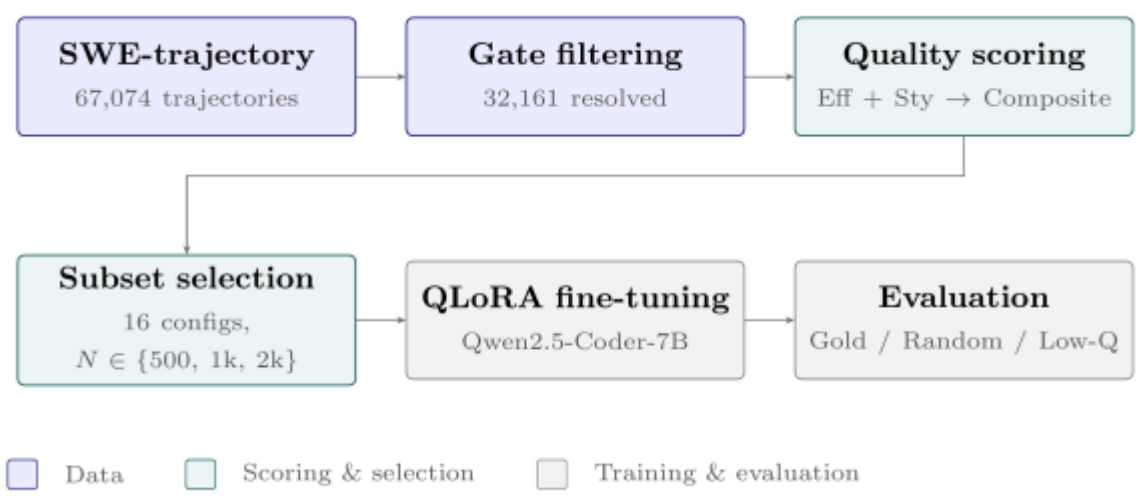

**Fig. 1.** Overview of the experimental pipeline

## 3.1 Composite Quality Scoring Framework

We decompose trajectory quality along two axes: Efficiency (whether the agent reaches the solution via a short, low-friction path) and Style (the regularity and diversity of action patterns). Correctness (resolved == 1) and Completeness (truncation ratio ≥ 0.90) are binary gate conditions rather than continuous scores; additionally, we require a parseable thought–action–observation structure (Format gate). The three gates avoid mixing binary and continuous variables in weighted aggregation.

After applying the three gates, 32,161 resolved trajectories remain (47.9% of 67,074). Within this pool we compute four sub-metrics. (i) B2 (Error-Retry): action → error → similar-action loops (same tool name; error keywords "Traceback", "Error", "failed"), capped at 10 and inverted so higher = better ($\sigma = 0.286$). (ii) B3 (Step-Count Ratio): step count over the median for the same task, clipped and inverted ($\sigma = 0.063$). Efficiency = mean(B2, B3). (iii) C2 (Action Diversity): normalized Shannon entropy over action types ($\sigma = 0.046$). (iv) C3 (Observation Utilization): fraction of observation entities (file basenames, error-class names) referenced in subsequent actions ($\sigma = 0.118$). Style = mean(C2, C3). Composite = 0.5 × Efficiency + 0.5 × Style ($\sigma = 0.083$, median 0.507). Equal weights are used as a neutral default to avoid post-hoc tuning; §4.5 confirms that dropping Style yields a negligible degradation ($\Delta = +0.0007$). Two candidates were excluded: B1 (Redundant Commands, $\sigma = 0.033$) and C1 (Observation Cleanliness, $\sigma = 0.043$), both ceiling-clustered (median > 0.96). B3 ($\sigma = 0.063$) is borderline; we retain it because it exhibits non-trivial dispersion, though Block 3 confirms it is the weakest retained sub-metric ($\Delta = +0.0016$).

### 3.2 Experimental Design

We design 16 fine-tuning experiments organized across three blocks plus an untrained baseline. Block 1 (Exp. 1–7, 14–15) compares Random, ResolvedOnly, TopQ, and BottomQ selection at 500, 1,000, and 2,000 trajectories. Block 2 (Exp. 8–9) ablates the high-level dimensions by ranking on Style only or Efficiency only. Block 3 (Exp. 10–13) ablates individual sub-metrics (NoB2, NoB3, NoC2, NoC3 at 500), and Exp. 16 retains only B2 as a single-metric baseline. All experiments use Qwen2.5-Coder-7B-Instruct [6] with identical LoRA hyper-parameters; trajectories are drawn from SWE-trajectory.

### 3.3 Evaluation Protocol

We evaluate on three test sets of 200 trajectories from a held-out partition: Gold (top 10% composite, primary), Random (uniform, excluding Gold), and Low-Q (bottom 10%). With ~47k tokens and ~64 turns per trajectory, each set contributes ~9.4M assistant tokens, providing substantially more signal for continuous loss estimation than a comparable count of binary SWE-bench outcomes. Three paradigms are used: (I) CE loss / perplexity over assistant tokens; (II) Next-Action Prediction at steps 3, 5, 7 with ROUGE-L and a JSON-format score; and (III) First-Action Evaluation, where the model is given the system prompt and issue description and must generate its first action, scored by action-type match, target-file match, and ROUGE-L. The expected ordering Gold < Random < Low-Q (H8) is a built-in framework check.

We adopt CE loss rather than end-to-end SWE-bench resolve rate for three reasons. (a) Statistical feasibility: 7B models attain ~1.0% resolve rate [15] on Verified, so detecting a 1-pp improvement at 80% power needs >3,000 instances per arm. (b) Scaffold confounding: end-to-end agent evaluation is highly sensitive to scaffold (prompt template, parser, temperature, ACI version) and can vary by 10+ percentage points; CE loss isolates training-data effects. (c) Measurement precision: CE loss is the SFT objective itself, computed over tens of thousands of tokens per trajectory. This follows established practice [9, 11, 13] and the scaling-laws literature [18], which establishes held-out CE loss as a capability proxy. We acknowledge that up-stream loss is not always strictly monotone with downstream task ability [19] and validate the proxy in Section 4.2 via first-action generation.

### 3.4 Training Configuration

All experiments use Qwen2.5-Coder-7B-Instruct [6] with QLoRA [14]. LoRA/quantization: rank r = 64, α = 128, dropout 0.05 on all seven linear projections; 4-bit NF4 + bfloat16; max context 32,768. Optimization: paged_adamw_8bit; LR $2 \times 10^{-5}$ with cosine decay (5% warm-up); 2 epochs; effective batch 8 (1 × 8 grad-accum); grad-clip 1.0; WD 0.01; gradient checkpointing + Flash Attention 2. Loss/format: tool calls use <tool_call> JSON tags; loss masks non-assistant tokens (−100). Compute: A100 80 GB, ≈150 GPU-hours. Training code, configurations, and scoring scripts are available at https://github.com/hanzunye/swe-trajectory-quality-study.

# 4 Experiments

## 4.1 Training Data Statistics

The Completeness Gate is vacuous on this dataset (truncation ratio = 1.0 for all 67,074 trajectories) but retained for portability. The Correctness Gate leaves 32,161 resolved trajectories (47.9%); we restrict scoring to these because unresolved trajectories carry erroneous intermediate actions—their effect is captured by the Random arms. Resolved trajectories average 64.4 assistant turns (range 11–100) and 47,171 tokens. Composite is approximately normal (mean = 0.518, $\sigma$ = 0.083). Sizes 500/1,000/2,000 span contemporary SWE-agent regimes (e.g., SWE-Gym [15] used <500), corresponding to the top 1.6%, 3.1%, and 6.2% of the resolved pool for composite-score selection.

## 4.2 Proxy Metric Validation

We validate CE loss via first-action generation. For four checkpoints (baseline, Random-500, TopQ-500, Random-1000) we prompt each model with the system instruction and issue description from 100 held-out Gold instances, generate up to 300 tokens, and compare to the ground-truth first-action via ROUGE-L and file-level match accuracy.

Table 1 reports the results. CE loss and ROUGE-L are perfectly rank-correlated (Spearman $\rho = -1.00$, $n = 4$). For $n = 4$ the minimum achievable two-sided Spearman p-value is $\approx 0.083$, so we do not claim $p < 0.001$; rather, we report this as the strongest possible monotonic alignment at this sample size and corroborate it with cross-paradigm consistency (perplexity vs. first action) and the qualitative behavioral evidence below. ROUGE-L improves from 0.137 (baseline) to 0.248 (Random-1000)—an 81% relative gain alongside a 54.5% CE-loss reduction. File-Match accuracy rises from 0.573 to 0.680 with fine-tuning, but does not strictly track CE loss among fine-tuned checkpoints, indicating that file targeting saturates earlier than token-level generation quality.

**Table 1.** Proxy metric validation via first-action generation (n = 100 instances per checkpoint, 300-token budget). CE Loss = mean cross-entropy on the 200-trajectory Gold set; ROUGE-L = LCS-based F-score against the ground-truth first action; File-Match = fraction of ground-truth target files referenced. Spearman $\rho$(CE Loss, ROUGE-L) = $-1.00$ ($n = 4$; minimum achievable two-sided $p \approx 0.083$).

| Strategy | CE Loss (Gold) | ROUGE-L | File-Match |
|---|---|---|---|
| No fine-tuning | 0.9100 | 0.137 | 0.573 |
| Random-500 | 0.4737 | 0.200 | 0.680 |
| Random-1000 | 0.4140 | 0.248 | 0.650 |
| TopQ-500 | 0.4704 | 0.212 | 0.670 |

Qualitatively, the not fine-tuned baseline emits structured markdown reports—no agentic behavior. After 500 trajectories the model adopts characteristic agent openings ("Let me start by understanding the problem…"); at 1,000 outputs are stylistically near identical to ground truth. CE-loss reduction thus reflects genuine capability acquisition rather than surface-level optimization. Action-Type Accuracy is 0.0 across the four validation checkpoints because the 300-token budget is fully consumed by the natural-language preamble before the <tool_call> tag is emitted—a measurement artefact rather than a capability claim; at 2,000 the same budget yields 4–7% Action-Type Accuracy (§4.4).

The two paradigms—perplexity and first-action generation—yield concordant conclusions across the three core hypotheses: SFT improves over baseline (CE 0.91→0.47, ROUGE-L 0.137→0.200); 500→1,000 scaling adds further gains (CE 0.474→0.414, ROUGE-L 0.200→0.248); and TopQ vs. Random at 500 are virtually indistinguishable (CE 0.470 vs. 0.474, ROUGE-L 0.212 vs. 0.200). With the qualitative shift, this establishes CE loss as a valid proxy for our comparisons.

**Table 2.** Cross-entropy loss (↓ lower is better) and perplexity on three test sets. ★ = best model. All models satisfy Gold < Random < Low-Q (H8 100% validated).

| Exp Nr. | Strategy | Gold ↓ | Random ↓ | Low-Q ↓ | Gold PPL |
|---|---|---|---|---|---|
| 0 | Baseline (no SFT) | 0.9100 | 0.9661 | 1.0812 | 2.484 |
| 1 | Random-500 | 0.4737 | 0.4875 | 0.5232 | 1.606 |
| 2 | Random-1000 | 0.4140 | 0.4266 | 0.4571 | 1.513 |
| 3 | TopQ-500 | 0.4704 | 0.4878 | 0.5260 | 1.601 |
| 4 | TopQ-1000 | 0.4106 | 0.4284 | 0.4624 | 1.508 |
| 5 | ResolvedOnly-500 | 0.4786 | 0.4886 | 0.5212 | 1.614 |
| 6 | ResolvedOnly-1000 | 0.4175 | 0.4260 | 0.4533 | 1.518 |
| 7 | BottomQ-500 | 0.4869 | 0.4916 | 0.5185 | 1.627 |
| 14 | Random-2000 | 0.3871 | 0.3939 | 0.4190 | 1.473 |
| 15 | TopQ-2000 ★ | 0.3732 | 0.3943 | 0.4269 | 1.452 |
| 16 | B2Only-500 | 0.4714 | 0.4868 | 0.5236 | 1.602 |

### 4.3 Block 1: Dataset Strategy Comparison

Table 2 reports CE loss across the 11 main configurations, including the 2,000-trajectory extension (Exp. 14–15) and the B2-only baseline (Exp. 16); ablation results (Exp. 8–13) appear in Table 4. All fine-tuned models substantially outperform the untrained baseline (Gold loss 0.910, PPL 2.484); the best, TopQ-2000, reaches Gold loss 0.3732 (PPL 1.452, 59.0% reduction). Every model satisfies Gold < Random < Low-Q (H8 100% validated).

Pairwise tests on per-trajectory Gold-set loss (two-sided Mann–Whitney U). H1 (Gate, L2): ResolvedOnly-500 (0.4786) does not improve over Random-500 (0.4737)—the difference runs counter to expectation ($\Delta = +0.0049$, $p = 0.78$). H2 (Score, L3): TopQ-500 (0.4704) is directionally better than ResolvedOnly-500 but not significant ($\Delta = -0.0082$, $p = 0.11$). H4 (Sanity, L3): TopQ-500 vs. BottomQ-500 is significant ($\Delta = -0.0165$, $p = 0.013$), confirming that the composite score captures meaningful quality variation.

**Table 3.** Scaling effect (500 → 1,000 → 2,000 samples) on Gold test set CE loss. All strategies show consistent improvement with diminishing but sustained returns. ResolvedOnly was not extended to 2,000 due to resource constraints.

| Strategy | Loss @500 | Loss @1000 | Loss @2000 | Δ 500→ 1000 | Δ 1000→ 2000 | Δ 500→ 2000 |
|---|---|---|---|---|---|---|
| Random | 0.4737 | 0.4140 | 0.3871 | −0.0597 (12.6%) | −0.0269 (6.5%) | −0.0866 (18.3%) |
| TopQ | 0.4704 | 0.4106 | 0.3732 | −0.0598 (12.7%) | −0.0374 (9.1%) | −0.0972 (20.7%) |
| ResolvedOnly | 0.4786 | 0.4175 | — | −0.0611 (12.8%) | — | — |

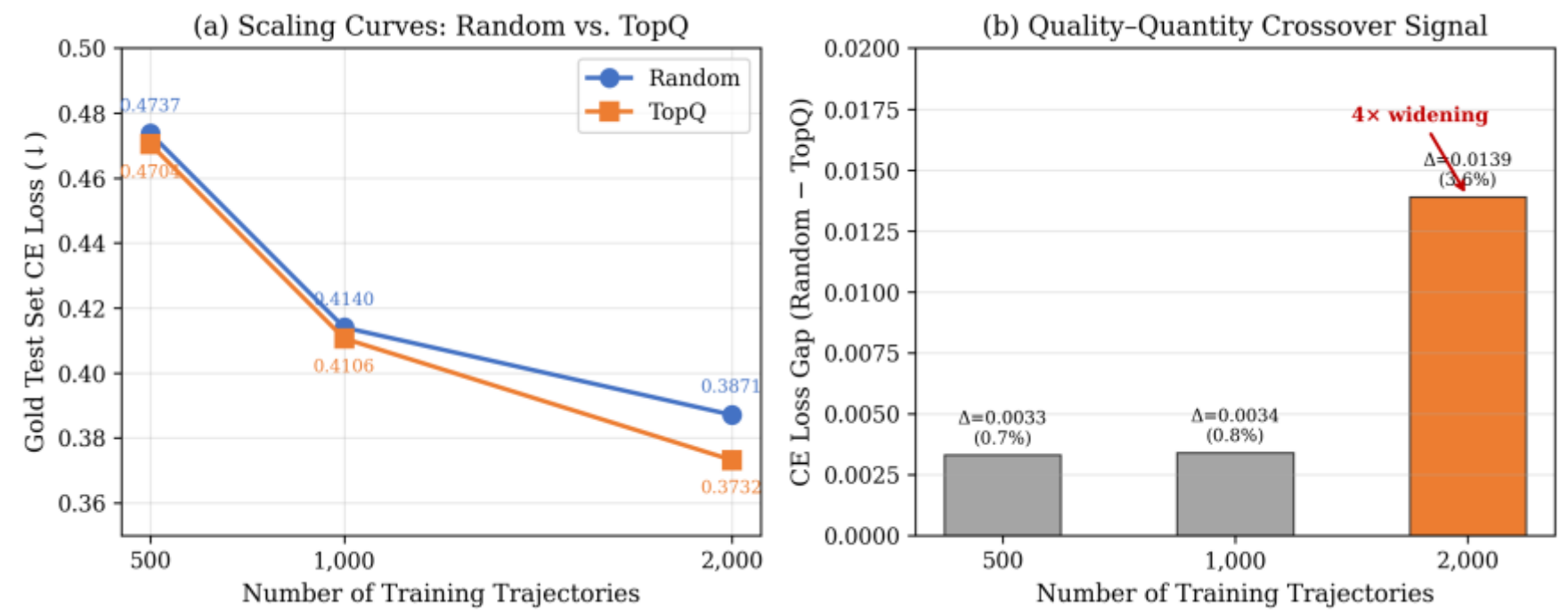

**Fig. 2.** (a) Random vs. TopQ on Gold-set CE loss at 500/1k/2k trajectories; gains diminish but persist. (b) The Random–TopQ gap widens 4× (0.7% → 3.6%), an emerging quality–quantity crossover signal.

## 4.4 Scaling Analysis: 500 → 2,000 Trajectories

H3 (Scaling, L1) is the most robust finding. Table 3 summarizes CE-loss reduction across 500/1,000/2,000 for Random and TopQ. Both deliver ~12.7% from 500→1,000, with smaller but sustained gains at 2,000 (Random −6.5%, TopQ −9.1%). At 500→1,000 the absolute drop ($\Delta \approx -0.060$) is ~18× the quality-filtering effect ($\Delta \approx -0.003$): in the small-to-medium regime, quantity dominates.

Extending to 2,000 reveals an emerging crossover signal (Table 3). At 500 TopQ (0.4704) and Random (0.4737) are indistinguishable ($\Delta = 0.0033$, 0.7%); at 1,000 the gap is similar ($\Delta = 0.0034$, 0.8%); at 2,000 it widens to $\Delta = 0.0139$ (3.6%), which a two-sided Mann–Whitney U test on per-trajectory Gold-set loss confirms as significant ($p = 0.016 < \alpha = 0.05$). TopQ's 1,000→2,000 improvement (−9.1%) also exceeds Random's (−6.5%), suggesting that quality filtering's marginal returns diminish more slowly. We treat this widening as a real signal of scale-dependent effective-ness, while a true crossover regime—where quality filtering's cumulative benefit matches a doubling of quantity—still requires direct experiments at ≥5,000 trajectories. At 2,000, first-action behavior also shifts qualitatively (33–36% of outputs contain structured edit or bash commands vs. <1% at 1,000), making ROUGE-L less informative at this scale.

**Table 4.** Combined ablation: high-level (Block 2), sub-metric (Block 3), and B2-only vs. composite at 500 samples. Δ and p are both relative to TopQ-500 (full composite); p from two-sided Mann–Whitney U on per-trajectory Gold-set CE loss (n = 200 per arm). All $p > 0.10$, consistent with the signal-to-noise argument in Section 5.1.

| Exp. | Configuration | Gold Loss | Δ vs. TopQ-500 | p-value |
|---|---|---|---|---|
| 3 | TopQ-500 (full composite, reference) | 0.4704 | — | — |
| 8 | NoEfficiency-500 (Style only) | 0.4774 | +0.0070 | 0.2666 |
| 9 | NoStyle-500 (Efficiency only) | 0.4711 | +0.0007 | 0.9108 |
| 10 | NoB2-500 (error-retry removed) | 0.4749 | +0.0045 | 0.4635 |
| 11 | NoB3-500 (step-count removed) | 0.4720 | +0.0016 | 0.7737 |
| 12 | NoC2-500 (action-diversity removed) | 0.4735 | +0.0031 | 0.5963 |
| 13 | NoC3-500 (obs-utilization removed) | 0.4700 | −0.0004 | 0.9273 |
| 16 | B2Only-500 (B2 alone, no composite) | 0.4714 | +0.0010 | 0.8780 |

## 4.5 Ablation: Dimensions, Sub-Metrics, and a B2-Only Baseline

We isolate each dimension and sub-metric (Table 4). Removing Efficiency (Exp. 8) yields the largest degradation versus the full composite ($\Delta = +0.0070$ vs. TopQ-500, $p = 0.27$); removing Style (Exp. 9) is essentially flat ($\Delta = +0.0007$, $p = 0.91$). The directional H5 contrast (NoEfficiency > NoStyle) gives a one-sided $p = 0.15$, confirming H5 directionally and indicating that Style contributes negligibly. At the sub-metric level, removing B2 (Exp. 10) is the largest Block-3 drop ($\Delta = +0.0045$ vs. TopQ-500, $p = 0.46$); the directional H6 contrast (NoB2 > NoB3) gives a one-sided $p = 0.31$, supporting H6 directionally. H7 is reversed (NoC3 $\Delta = -0.0004$ vs. NoC2 $\Delta = +0.0031$), suggesting C3 may inject noise. None of the Block-2/3 ablations reaches significance versus TopQ-500 (all two-sided $p > 0.25$). Composite-ranked TopQ-500 vs. B2-only-ranked TopQ-500 (Exp. 16) differs by < 0.2% on Gold CE loss ($\Delta = +0.0010$, $p = 0.88$), with first-action ROUGE-L slightly favoring composite (0.212 vs. 0.198) and File-Match favoring B2-only (0.690 vs. 0.670). At 500, B2 alone captures the dominant signal; the composite's value may grow at scale, consistent with the widening Section 4.4 gap.

# 5 Analysis and Discussion

## 5.1 Interpretation

The quality null results (H1, H2, H5–H7) follow a signal-to-noise argument: the within-group σ of Gold loss (≈ 0.07) far exceeds the maximum 500-sample effect ($\Delta \approx 0.016$, TopQ vs. BottomQ), so the per-arm size required for 80% power lies far above our 200. By contrast, the scaling effect dominates—12.7% per doubling at 500→1,000 with sustained 6.5–9.1% returns at 2,000—and the TopQ–Random gap widens from 0.7% to 3.6%, consistent with quality filtering becoming increasingly relevant at scale; a precise crossover location requires substantially larger scales. The primacy of B2 is intuitive: high-retry trajectories add noisy supervision via the error-correction loop, whereas low-retry trajectories more directly encode the state→action mapping.

## 5.2 Tool-Call Format Failure

Tool-call JSON format compliance (i.e., emission of well-formed <tool_call> tags) is zero across all 16 fine-tuned configurations regardless of selection strategy, scale, or scoring approach. The cause is class imbalance: bash and edit actions dominate assistant turns (11–68% and 32–86%) while tool_call accounts for only 0–3%, leaving the LoRA adapter little exposure to well-formed JSON. Deployment must therefore use constrained decoding or targeted format-aware augmentation.

## 5.3 Comparison with General-Purpose Data Selection

Adapting selectors such as IFD [13] and DEITA [11] to trajectories faces three structural mismatches: (i) multi-step quality is multi-dimensional; (ii) each action depends on preceding observations; and (iii) action spaces differ from natural-language responses. A naive treatment of the trajectory as a single response would conflate per-action quality with trajectory coherence. Our framework targets these features, and the TopQ–BottomQ sanity check ($p = 0.013$) confirms it captures meaningful variation. Trajectory-native adaptations of IFD/DEITA remain future work.

## 5.4 Limitations

(1) Statistical power & crossover scope: each test split has 200 trajectories; ~2,800 are needed to detect a 0.7%-magnitude effect at 80% power. The scaling effect (~12.7%) is well above this bound, while quality-filtering effects at 500–1,000 lie below it; the 3.6% gap at 2,000 reaches significance (Mann–Whitney $p = 0.016$), but any specific crossover location remains bounded by the three scale points and the 200-trajectory split. (2) Proxy scope: the CE-loss/ROUGE-L alignment is over $n = 4$ checkpoints (minimum two-sided Spearman $p \approx 0.083$), so the alignment is supporting rather than conclusive. (3) Generalization: a single base model and trajectory source; replication on other code LLMs (e.g., Code Llama [7], DeepSeek-Coder [8]) or trajectory corpora (SWE-smith [16] / R2E-Gym [17]) is open. (4) Evaluation scope: a 300-token generation budget yields Action-Type Accuracy = 0 at 500–1,000 and compresses ROUGE-L at 2,000; sensitivity to longer budgets is not characterized, and IFD/DEITA base-lines are not included for the structural reasons in Section 5.3.

## 6 Conclusion

We presented a systematic study of data filtering for LoRA fine-tuning of a code-agent LLM with 16 controlled experiments across three scales (500/1,000/2,000). Methodologically, we introduce a multi-dimensional composite quality score (Efficiency: B2, B3; Style: C2, C3) for code agent trajectories beyond binary pass/fail filtering. The framework is self-consistent (TopQ vs. BottomQ, $p = 0.013$), and ablation identifies B2 as the dominant sub-dimension—B2 alone matches the full composite at 500 ($\Delta < 0.2\%$). We also validate CE loss as a proxy via first-action generation (Spearman $\rho = -1.00$, $n = 4$).

Empirically, in this regime data quantity dominates quality filtering: doubling 500→1,000 yields ~12.7% relative loss reduction; quality filtering yields <1% (n.s.). At 2,000 the TopQ–Random gap widens to 3.6% (Mann–Whitney $p = 0.016$), and TopQ's 1,000→2,000 improvement (−9.1%) exceeds Random's (−6.5%)—a real signal of scale-dependent effectiveness rather than a confirmed crossover.

These conclusions are specific to Qwen2.5-Coder-7B-Instruct on SWE-trajectory; replication on other models and corpora, direct experiments at ≥5,000 trajectories, trajectory-native IFD/DEITA, and the tool-call format failure are the most important next steps.